\pdfoutput=1

\documentclass[11pt]{article}

\usepackage[preprint]{acl}

\usepackage{times}
\usepackage{latexsym}

\usepackage[T1]{fontenc}

\usepackage[utf8]{inputenc}

\usepackage{microtype}

\usepackage{inconsolata}

\usepackage{graphicx}
\usepackage{lipsum}
\usepackage{svg}
\usepackage{booktabs}
\usepackage{highlight}
\usepackage{amsmath}
\title{WangLab at MEDIQA-M3G 2024: Multimodal Medical Answer Generation using Large Language Models}

\author{
Ronald Xie$^{1,2}$ \quad Steven Palayew$^{1,2}$ \quad Augustin Toma$^{1,2}$ \quad Gary Bader$^{1,3,4}$ \quad Bo Wang$^{1,2,5,6}$\vspace{5pt} \\
$^{1}$University of Toronto \quad
$^{2}$Vector Institute \quad 
$^{3}$Princess Margaret Cancer Centre, University Health Network \quad \\
$^{4}$Lunenfeld-Tanenbaum Research Institute, Sinai Health System \quad \\
$^{5}$Peter Munk Cardiac Centre, University Health Network \quad 
$^{6}$AI Hub, University Health Network\vspace{5pt}\\
{\footnotesize \texttt{\{augustin.toma, ronald.xie, steven.palayew\}@mail.utoronto.ca}, gary.bader@utoronto.ca, bowang@vectorinstitute.ai}
}

\begin{document}
\maketitle

\begin{abstract}
This paper outlines our submission to the MEDIQA2024 Multilingual and Multimodal Medical Answer Generation (M3G) shared task. We report results for two standalone solutions under the English category of the task, the first involving two consecutive API calls to the Claude 3 Opus API and the second involving training an image-disease label joint embedding in the style of CLIP for image classification. These two solutions scored 1st and 2nd place respectively on the competition leaderboard, substantially outperforming the next best solution. Additionally, we discuss insights gained from post-competition experiments. While the performance of these two solutions have significant room for improvement due to the difficulty of the shared task and the challenging nature of medical visual question answering in general, we identify the multi-stage LLM approach and the CLIP image classification approach as promising avenues for further investigation.

\end{abstract}

\section{Introduction}
An increased demand for healthcare services and recent pandemic needs have accelerated the adoption of telehealth, which was previously underused and understudied \citep{Shaver2022-ny, mediqa-m3g-task}. There has been significant recent interest in integrating artificial intelligence (AI) into telehealth \citealp{ma2024segment,toma2023clinical}, as these technologies have the potential to enhance and expand its ability to address important healthcare needs \citep{Sharma2023-bj}. The task of consumer health question answering, an important part of telehealth, has been explored actively in research. However, the focus of this existing research has been on text \citep{Ben_Abacha2019-xw}, which is limiting as medicine is inherently multimodal in nature, requiring clinicians to work not just with text but also with imaging among other modalities \citep{Corrado2023-cm}. 

To help address this gap, the MEDIQA-M3G shared task was proposed \citep{mediqa-m3g-task}. This task requires the automatic generation of clinical responses given relevant user generated text and images as input, with a specific focus on clinical dermatology \citep{mediqa-m3g-task}. 

This work describes our submission to this task. We explored two standalone solutions, one involving two consecutive API calls to the recently released Claude 3 Opus model \citep{Anthropic_undated-gg} and the other trains a joint image-disease label embedding model using CLIP \citep{Radford2021-bd} for image classification. These two strategies took 1st and 2nd place respectively during the competition. While our strategy’s effectiveness relative to other submissions highlight that Claude 3 Opus and multi-stage LLM frameworks have potential value in the area of multi-modal medical AI, both our solutions' performance is limited despite their leaderboard success, highlighting the difficulty of the shared task and the unsolved challenge of medical visual question answering. 

\section{Shared task and provided dataset}

The MEDIQA-M3G competition focuses on the problem of clinical dermatology multimodal query response generation. The inputs include text which give clinical context and queries, as well as one or more images associated with the case \citep{mediqa-m3g-dataset}. The task is to generate responses to these cases resembling those made by medical professionals in the field of dermatology. Participants have the option to generate these responses in three languages: Chinese (Simplified), English, and Spanish. \citep{mediqa-m3g-task} 

The dataset consists of 842 train, 56 validation, and 100 test cases. Each case consists of one or more images of skin conditions, their accompanying query text which may or may not include clinical context, patient queries, additional details regarding the disease and in some cases possible diagnosis. Finally, for each case there are multiple responses made by one or more medical professionals, which are used as targets to score the model predictions. The cases also notably include metadata on the rank and validation level of the authors of content, which are used in evaluation \citep{mediqa-m3g-dataset}. For evaluation, the competition uses a version of the deltaBLEU \citep{Galley2015-rf} metric to allow a single score to be computed based on word matching, weighted by the consistency (most frequent response) and the seniority of the medical professional across all responses given for that particular case. \citep{mediqa-m3g-task}

The query text and target responses are given in multiple languages, namely Chinese, English, and Spanish \citep{mediqa-m3g-dataset}. It's worth noting that while the test and validation sets were translated by medical professionals, the training set of 842 cases seems to be translated automatically with some potential room for errors. For our submission we focus on only providing the English solution.

\section{Related Work}
 
There has recently been a substantial amount of interest in medical applications of multimodal machine learning, and large multimodal models. Some notable examples of research in this area include the open source LLAVA-MED model \citep{Li2023-mj}, and ELIXR, with the latter, similar to our work, exploring not only the application of large multimodal models, but also training a model using CLIP \citep{Xu2023-yf}. However, while there has been significant focus on certain areas such as radiology, the area of dermatology has not been explored to the same extent. Cirone et al. notably found that GPT-4V could accurately differentiate between benign lesions and melanoma \citep{Cirone2024-al}. However, this is a much less challenging task than the one proposed in this shared task, as the problem space is much smaller in scope than responding to dermatology questions which are not necessarily in the train set, with even the conditions of interest not necessarily being in the train set. The limited performance of our solution, along with it being by far the best performing solution in this competition demonstrate the challenge of this task, and highlight the need for significant progress before deployment in a clinical setting. However, our work highlights potentially important directions for future research, including further investigation of multi-stage LLM systems, and the importance of evaluation metrics in the benchmarking of the clinical efficacy of developed systems.


\begin{figure*}[t]
\centering
\includegraphics[width=\textwidth]{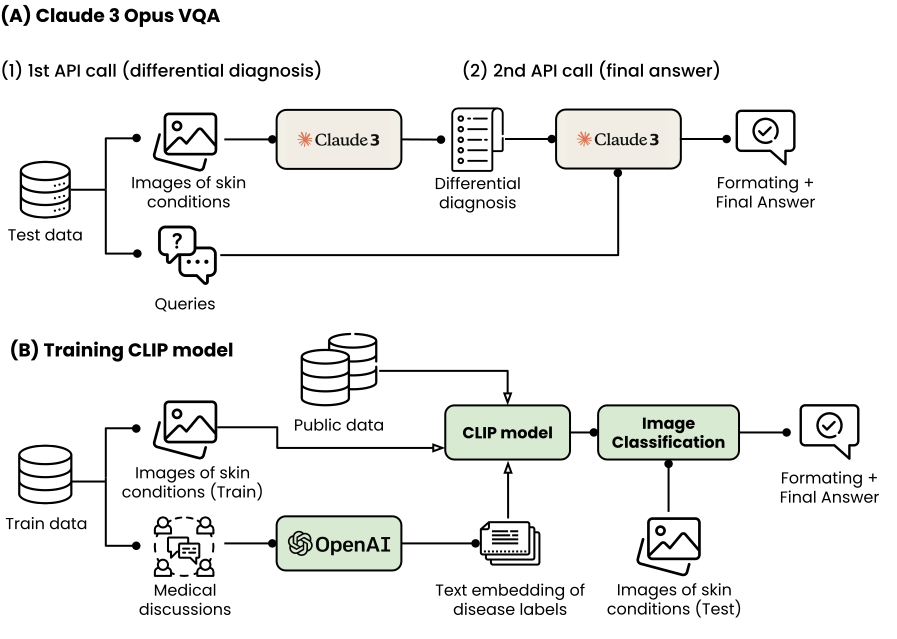}
\caption{Overview of the two winning solutions. A) Test cases are directly submitted to the Claude 3 Opus API. The first of the two consecutive API calls generates differential diagnosis using only the images in the test cases and the second API call optionally includes the associated queries, specifies formatting, and generates final answer. B) The medical discussions included as a part of the training data is used to extract the most likely disease label for each case using GPT4-Turbo from OpenAI. The resulting image-disease label pair are used in conjunction with publicly available data to train a joint embedding in the style of CLIP. The disease labels are embedded using EmbeddingV3 from OpenAI and used to train the image encoder (ResNet50) and both the image and text projection layers. Finally, once the model is trained, the test images are classified inside the learned joint embedding which becomes the final output before performing post processing. }
\label{fig:overview}
\end{figure*}

\section{Results}

Upon examination of the evaluation metric and competition data, we have determined that a short response focusing on disease diagnosis alone is the most advantageous. This is due to two reasons. First, we notice both the training and validation sets often contain short responses, and in many cases merely the skin condition presented in the associated images. Second, the evaluation metric's penalty for short responses is significantly smaller than a longer, partially correct response. Given these initial findings, we evaluated two methods as outlined in \ref{fig:overview} which took 1st and 2nd place in the English category of the leaderboard during the MEDIQA-M3G challenge by a significant margin over the next best submitted solution, the latter of which received a deltaBLEU score of 3.827 during the competition. The methods will be elaborated in the following sections in detail. 

\subsection{Claude 3 Opus API solution} 
The higher scoring of the two methods consists of two successive API calls to Claude 3 Opus \citep{Anthropic_undated-gg}. For each case in the test set, the first API call generates possible differential diagnosis for the given images, and the second API call further processes the response into the name of the most likely disease only, which is then returned. 

This exact configuration was decided based on trial and error. Table \ref{claude-results} outlines the solutions tested. Notably, we observe that the disease diagnosis given by Claude 3 Opus was poorer quality when the prompt constrains the output format upon manual review. This was further confirmed by the inferior performance of the 1-call result. Therefore, we let the API generate differential responses only with the provided images alone without any constraints on format, and use a second API call to reformat the response into the desired form, which is just the name of the skin condition without any abbreviations. 

Furthermore, we observe that including the accompanying query text for each case either in the 1st or 2nd pass was not able to outperform simply using the image alone to make the predictions. This finding may be attributed to the inconsistent information present in the query text, which may often harm the prediction from Claude 3 in some cases. It may also be a potential limitation of Claude's ability to reason with text and image simultaneously. Indeed the resulting predictions had substantial room for improvement even under the most favorable setting tested. All prompts used to produce the solutions in Table \ref{claude-results}, including the winning solution are outlined in Appendix.

\begin{table*}[t] 
\noindent
\centering
\caption{Performance of various Claude 3 Opus based solutions. 1 Call involves simply generating a response based on images, whereas 2 Calls involve first generating a differential diagnosis, then using a second API call to come up with a final diagnosis. For Img+text, both modalities are used in the first API call to generate the differential, whereas for Img then text the first API call uses only images, then the second API call uses text}
\label{claude-results}
\setlength\tabcolsep{4pt} 
\begin{tabular}{@{}lcccccc@{}}
\toprule
Scen. & dBLEU & BP & Ratio & Hyp\_len & Ref\_len \\
\midrule
Img (1 Call) & 7.650 & 0.984 & 0.984 & 498 & 506 \\
Img (2 Calls) & 10.415 & 0.994 & 0.994 & 485 & 488 \\
Img then text (2 Calls) & 8.775 & 0.983 & 0.983 & 527 & 536 \\
Img + text (2 Calls) & 8.803 & 1.000 & 1.004 & 523 & 521 \\
\bottomrule
\end{tabular}
\end{table*}

\subsection{CLIP image classification solution}

\subsubsection{Image classification via nearest neighbour retrieval}
Once the image encoder and the respective image and text projection layers are trained, the resulting joint embedding can be used to perform image classification via nearest neighbour retrieval. Specifically, we embed each image associated with a given case in the competition test set and find 5 nearest neighbours for each embeded image. We test 4 different conditions, namely retrieval between the image embedding of the query (testing dataset) and either its nearest 5 text or image embeddings from the reference (training dataset), and whether the nearest neighbours are computed in PCA space or as normal. We then pool the labels associated with the retrieved examples and return the most frequent as the final predicted label for the case. The resulting scores are presented in Table \ref{tab:clip-retrieval}. Of note, during the competition, random augmentations were mistakenly not turned off during inference when obtaining the image embeddings. This did not lead to better performance and was corrected after the competition concluded. 

\subsubsection{Importance of batch size}
The CLIP loss heavily relies on a diverse source of positive and negative pairs to converge to a good solution. It's often the case that bigger batch sizes give more robust joint representations. However, under low data settings such as for this competition where the available labelled data is scarce, larger batch sizes may lead to overfitting which is destructive for generalization. We test 3 different batch sizes ranging from 128 to 512 and observe that a batch size of 256 is most suitable for the task and the amount of training data available. The results are presented in Table \ref{tab:clip-batchsize}. 

\begin{table}[ht]
\centering
\caption{Performance of the CLIP based solution across different batch sizes}
\label{tab:clip-batchsize}
\resizebox{\columnwidth}{!}{%
\begin{tabular}{lccccc}
\toprule
Model & dBLEU & BP & Ratio & Hyp\_len & Ref\_len \\
\midrule
CLIP (batch 128) & 7.848 & 0.980 & 0.980 & 483 & 493 \\
CLIP (batch 256) & 8.404 & 0.966 & 0.966 & 461 & 477 \\
CLIP (batch 512) & 8.187 & 0.983 & 0.984 & 447 & 485 \\
\bottomrule
\end{tabular}%
}
\end{table}

\begin{table}[ht]
\centering
\caption{Performance of CLIP with different retrieval related strategies, including retrieval in the PCA space, and retrieving based on either the image or the text embedding of the reference. Of note, the random image augmentations during inference were enabled unintentionally during the competition but disabled for all subsequent experiments. }
\label{tab:clip-retrieval}
\resizebox{\columnwidth}{!}{%
\begin{tabular}{cccll}
\toprule
\textbf{Random. Aug} & \textbf{PCA Space} & \textbf{Query-Reference} & \textbf{dBLEU} \\
\midrule
Yes & Yes & Image-Image & 8.744 \\
No & No & Image-Text & 9.262 \\
No & Yes & Image-Text & 6.279 \\
No & No & Image-Image & 10.119 \\
No & Yes & Image-Image & 8.404 \\
\bottomrule
\end{tabular}%
}
\end{table}

\subsection{{Post processing}}
Post processing is performed on both the Claude 3 Opus API solution and the CLIP based image classification solution in the same way. This includes putting the output disease name in predetermined sentence format to mimic the style of the given responses from medical professionals, specifically in the form of "It is [Disease name].". While a naive approach to the VQA task, we find this simple formatting allows our disease labels produced from images alone to score quite competitively under the deltaBLEU evaluation metric provided by the competition organizers compared to simply returning the disease name itself. 

Furthermore, unlike other competitors' solutions based on finetuning existing VQA models (such as LLaVA-med) simultaneously using both the images and the associated query text, our solution does not take advantage of any potentially useful information included in the query text. As a naive way of overcoming this limitation, we simply compiled a dictionary of disease names present in the training data and do simple word matching with the query text and replaced any matches to the query text as the disease condition. While this often times does not produce the correct diagnosis, considering the difficulty of the task this approach does confer some improvement in overall deltaBLEU score. The ablations of the post processing is outlined in Table \ref{ablation-processing}.

\begin{table}[ht]
\centering
\resizebox{\columnwidth}{!}{%
\begin{tabular}{lccc}
\hline
Solution & Word Matching & Sentence Structure & Both \\
\hline
Claude Solution & 3.580 & 5.741 & 10.415 \\
CLIP Solution (competition) & 2.452 & 2.041 & 8.744 \\
CLIP Solution (batch 256) & 3.334 & 5.092 & 10.119 \\
\hline
\end{tabular}
}
\caption{Result of ablations on performance of top performing solutions. Sentence structure involves placing the predicted disease labels in predetermined sentence format, whereas word matching is a heuristic employed to utilize provided text via naively matching disease names with the given queries.}
\label{ablation-processing}
\end{table}

\begin{table}[ht]
\centering
\begin{tabular}{lc}
\hline
\textbf{HyperParameter}             & \textbf{Value}                                    \\ \hline
Image encoder                  & Resnet50                                          \\
Projection dim                 & 256                                               \\
Batch size                     & 256                                               \\
Text embedding dim             & 3072                                              \\
Image embedding dim            & 2048                                              \\
Num. projection layers         & 1                                                 \\
Augmentations                  & \begin{tabular}[c]{@{}c@{}}RandFlip, RandRotate,\\ RandSpatialCrop,\\ RandAdjustContrast\end{tabular} \\
Weight decay                   & 0.001                                             \\
Learning rate                  & 0.001                                             \\ \hline
\end{tabular}
\caption{Hyperparameters corresponding to the highest performing CLIP solution}
\label{tab:clip-hyperparameters}
\end{table}

\section{Discussion}
We have presented two solutions to the MEDIQA2024-M3G competition, one involving API calls to an existing state of the art multimodal language model and the other involving the learning of an image-disease label joint embedding space for disease classification. 

The superior performance of using two separate API calls to Claude 3 Opus over one pass was interesting to observe. The increase in performance is likely attributed to the reduced ability for the model to simultaneously reason with the images while adhering to the added difficulty of only returning the disease label without any additional textual generation. This finding is somewhat consistent with how chain of thought reasoning can improve model performance by asking the model to first consolidate evidence present in the given image followed by making several differential diagnoses. Further research such as \citep{Zhang2023-ci} also highlight the importance of using two-stage frameworks for multi-modal chain of thought that separate rationale generation and answer inference over one stage systems. 

For the CLIP based solution, our additional experiments after the competition highlights the importance of proper selection of batch size and retrieval method. Effective convergence of the CLIP loss hinges on a rich set of positive and negative training pairs. While larger batch sizes generally yield more robust joint representations, limited data settings, as encountered in the case for this competition, can suffer from overfitting with larger batches. This overfitting hinders the model's ability to generalize. Furthermore, we explored neighbour retrieval settings for image classification during testing. We observe that while CLIP effectively constructs a joint embedding space between images and their disease labels, the image embeddings and text embeddings remain as separate cluster in PCA space. As a result, we see that the nearest 5 neighbours in the text cluster for each embeded image (image-text) in the test set were much poorer in quality than those retrieved from the image cluster (image-image). 

\section{Limitations}
While both the Claude 3 Opus API based solution and the CLIP based image classification solution achieved first and second place during the MEDIQA-M3G competition respectively, they have substantial room for improvement despite their leaderboard success.  

First of all, the overall deltaBLEU score of both solutions are poor, hovering just above 10. The scores during the competition were also unable to be reproduced given the provided evaluation code, which produced systematically lower scores compared to those received during the competition despite the same solution being used during evaluation in both cases. Nevertheless, the low absolute scores of the solutions really highlight the difficulty of the medical VQA task presented and the difficulty of such tasks in general. Upon examining the solutions, we observe that the models were seldom able to generate the exact name of the skin condition in question, although do a good job at identifying a disease similar in presentation or effect location (for example tinea scalp vs seborrheic dermatitis). Certainly both solutions require substantial improvements before they contribute meaningful benefits to the healthcare system in practice. 

Next, both solutions while reproducible are not stable. The Claude API may be subject to randomness during generation due to the temperature parameter or the update of internal private model weights while the CLIP solutions observed inconsistencies during retrieval where the retrieved images seldom agreed with each other, leading to low confidence in the final output. Retraining the CLIP model with the same experimental setup but initializing differently may yield completely different final disease label classification due to this inconsistency. 

Lastly, the two solutions were formulated with the competition evaluation metric in mind as they are both framed as a disease label prediction task rather than a more traditional VQA task which could cover topics such as differential diagnoses, treatments and other recommendations as present in the actual medical discussions which were treated as ground truths for this competition. This is further reason to treat the performance of the presented solutions with a grain of salt. Specifically, upon our initial exploration, the deltaBLEU metric defined by the competition organizers favors short responses given the relatively heavy penalty incurred on incorrect k-mers present and relatively low penalty on a incomplete answer in comparison. This discourages model exploration during text generation and potentially penalizes model predictions that are correct semantically but are either too long or not containing the exact words present in the ground truths. This is highlighted in the ablation results in Table \ref{ablation-processing}. Furthermore, the naive word matching often gave incorrect diagnosis as the patient writing the query does not have medical background, however the solution containing the disease label still scored well under the current metric as medical professionals respond with "not [disease label]" which has the opposite semantic meaning but similar k-mer composition. We recommend the organizers to slightly modify the existing metric to be more lenient with assessing the produced solutions and perhaps add a semantics component in addition to a k-mer based evaluation metric such as GPTscore \citep{Fu2023-ee}, that can provide more robustness in assessing the quality of generated responses.

Nevertheless, the competition still serve as an important step towards the goal of automatically generating clinical responses given textual queries and associated images, and we sincerely thank the organizers for the work putting together this dataset and for hosting the competition.

\section{Conclusion}

We present two solutions to the English category of the MEDIQA2024-M3G shared task for Multilingual and Multimodal Medical Answer Generation. The Claude 3 Opus API based solution and the CLIP image classification based solution scored 1st and 2nd, respectively among all submissions. While there is still substantial room for improvement for these two solutions, we share and discuss our findings to contribute towards the important goal of automatically generating clinical responses given textual queries and associated images.



\clearpage

\bibliography{custom}

\clearpage

\appendix
\setcounter{figure}{0} 
\renewcommand{\thefigure}{S\arabic{figure}} 

\section{Claude 3 Opus API prompts}
Example prompts used to perform API calling in the Claude 3 Opus solution and other tested variants. 
\subsection{Image only 1-call}
\textbf{System:} You are an expert assistant to a blind dermatology student, help him identify exactly what conditions would be included in the differential for this condition? Be concise. After brief description of the images and explanation of your choice, give the most commonly occuring skin disease out of the differentials at the end and nothing else, in the form of \\\\Answer: [Disease Name]\\\\ \\
\textbf{Content:} IMG\_ENC00908\_00001.jpg, IMG\_ENC00908\_00002.jpg \\
\textbf{Output:} Answer: Dyshidrotic eczema

\subsection{Image only 2-calls}
\textbf{System:} You are an expert assistant to a dermatology student, help him identify exactly what skin conditions would be included in the differential for the images presented. Consider both resemblence and prevalence. \\
\textbf{Content:} IMG\_ENC00908\_00001.jpg, IMG\_ENC00908\_00002.jpg \\
\textbf{Output1:} Based on the images provided, the key skin findings are ... The differential diagnosis for these lesions would include:\\\\1. Hand eczema (dyshidrotic eczema) ... \\
\textbf{System:} You are an expert assistant to a dermatology student. Given the following differentials, only return the name of the most likely diagnosis and nothing else. Do not include alternative names of the differential in brackets. \\
\textbf{Content:} Based on the images provided, the key skin findings are ... The differential diagnosis for these lesions would include:\\\\1. Hand eczema (dyshidrotic eczema) ... \\
\textbf{Output2: } hand eczema

\subsection{Image then text 2-calls}
Of note, the first API remains the same to the Image only 2-calls case, but the added Additional Information field contains the text query associated with each case in the test set. 

\textbf{System:} You are an expert assistant to a dermatology student, help him identify exactly what skin conditions would be included in the differential for the images presented. Consider both resemblence and prevalence. \newline
\textbf{Content: }IMG\_ENC00908\_00001.jpg, IMG\_ENC00908\_00002.jpg \newline
\textbf{Output1: } Based on the images provided, the key skin findings are … The differential diagnosis for these lesions would include:\\\\1. Hand eczema (dyshidrotic eczema) … \newline
\textbf{System:} You are an expert assistant to a dermatology student, given the following differentials discussed and some additional information provided, only return the name of the most likely diagnosis and nothing else. Do not include alternative names of the differential in brackets.  \newline
textbf{Content: } Differentials: \\\\ Based on the images provided, the key skin findings are … The differential diagnosis for these lesions would include:\\\\1. Hand eczema (dyshidrotic eczema) … Additional information: Picture 1:  On the outside of the thigh, there is a small circle of lump.  Approximately 2 months.\\Picture 2:  Small red spots on the palm.  There is slight numbness in the palm.
\textbf{Output2: } hand eczema (dyshidrotic eczema)

\subsection{Image + text 2-calls}

\textbf{System:} You are an expert assistant to a dermatology student, help him identify what skin conditions would be included in the differential for the presented images and additional information provided by the medical professional. If any skin conditions are mentioned in the additional information, include them as the most likely differential. \newline
\textbf{Content: }Additional information: Picture 1:  On the outside of the thigh, there is a small circle of lump.  Approximately 2 months.\\Picture 2:  Small red spots on the palm.  There is slight numbness in the palm. \newline IMG\_ENC00908\_00001.jpg, IMG\_ENC00908\_00002.jpg \newline
\textbf{Output1: } Based on the provided images and additional information, here are the potential skin conditions to consider in the differential diagnosis: ...
\textbf{System:} You are an expert assistant to a dermatology student. Given the following differentials, only return the name of the most likely diagnosis and nothing else. Do not include alternative names of the differential in brackets. \newline
textbf{Content: } Based on the provided images and additional information, here are the potential skin conditions to consider in the differential diagnosis: ... \newline
\textbf{Output2: } picture 1: lipoma. picture 2: palmar erythema

\begin{figure*}[ht]
\centering
\includegraphics[width=1.5\columnwidth]{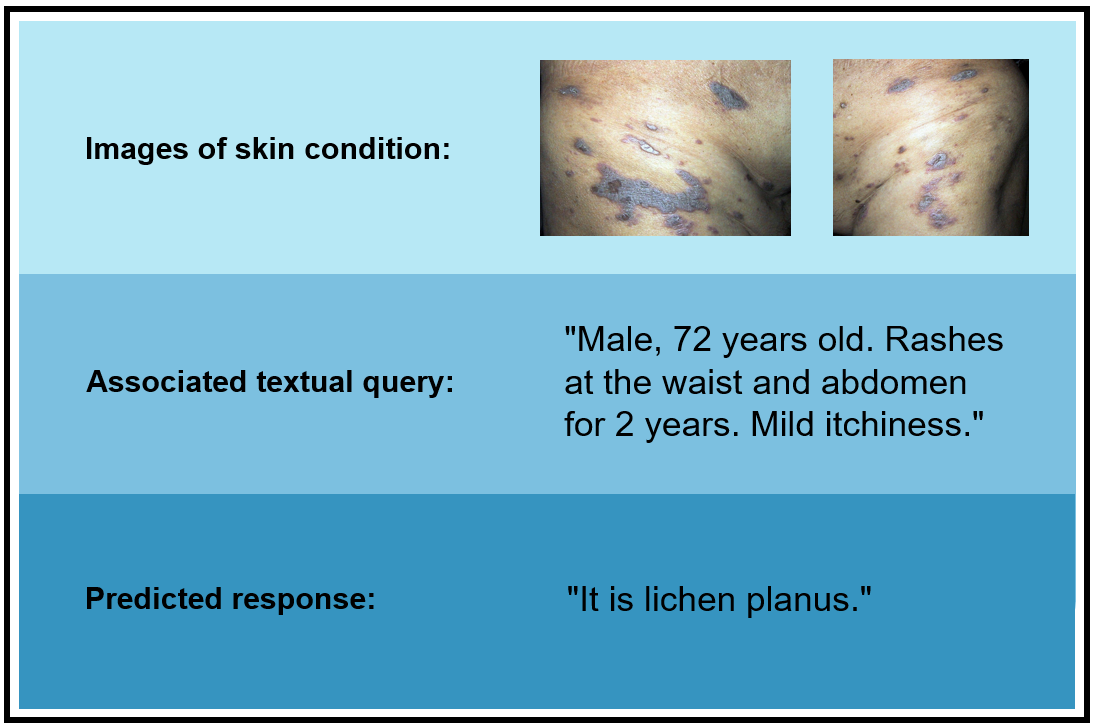}
\caption[]{Representative case example illustrating the images of the skin condition, their associated textual query and the predicted response given.}
\label{appendix-case-example}
\end{figure*}

\begin{figure*}[ht]
\centering
\includegraphics[width=\textwidth]{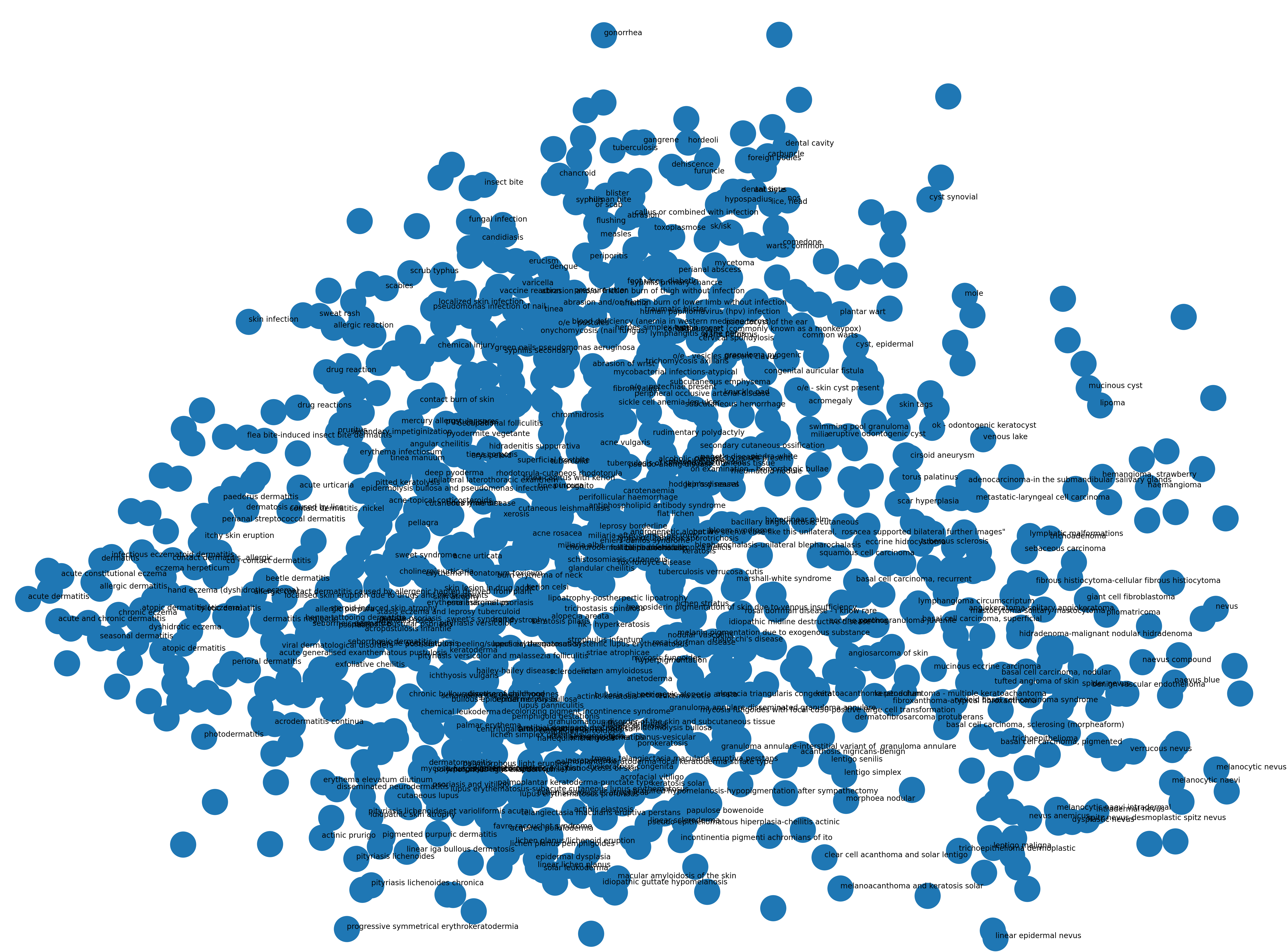}
\caption[]{PCA visualization of all the training disease labels embedded by the EmbeddingV3 model. Skin conditions that are semantically similar are clustered together in this representation space.}
\label{appendix-PCA-embedding}
\end{figure*}

\end{document}